\documentclass[10pt,twocolumn,letterpaper]{article}

\usepackage[pagenumbers]{cvpr} 
\usepackage{hyphenat}

\usepackage{booktabs, siunitx}
\usepackage{comment}
\usepackage{float}
\usepackage{multirow, xcolor, colortbl}
\usepackage{pifont}
\usepackage{footmisc}

\newcommand{\clsicon}[1]{\textcolor{#1}{\rule{0.6em}{0.6em}}}

\newcommand{\cityclsicon}[1]{\textcolor{#1}{\rule{0.7em}{0.7em}}}

\definecolor{mgray}{HTML}{E6E6E6}
\definecolor{road}{rgb}{0.651, 0.325, 0.651}
\definecolor{otherstruct}{rgb}{0.522, 0.522, 0.796}
\definecolor{pole}{rgb}{0.780, 0.780, 0.780}
\definecolor{twowheeler}{rgb}{0.000, 0.000, 1.000}
\definecolor{truck}{rgb}{0.000, 0.000, 0.357}
\definecolor{car}{rgb}{0.000, 0.000, 0.725}
\definecolor{pedestrian}{rgb}{1.000, 0.102, 0.306}
\definecolor{terrain}{rgb}{0.776, 1.000, 0.776}
\definecolor{vegetation}{rgb}{0.545, 0.725, 0.180}

\makeatletter
\newcommand{\smallscriptsize}{\@setfontsize\smallscriptsize{8.5}{10}} 
\makeatother
\usepackage{float}
\definecolor{cvprblue}{rgb}{0.21,0.49,0.74}
\usepackage[pagebackref,breaklinks,colorlinks,allcolors=cvprblue]{hyperref}

\def\paperID{21196} 
\def\confName{CVPR}
\def\confYear{2026}

\title{MobileOcc: A Human-Aware Semantic Occupancy Dataset for Mobile Robots}

\author{%
\parbox{\textwidth}{\centering
Junseo Kim\textsuperscript{*} \quad
Guido Dumont\textsuperscript{*} \quad
Xinyu Gao\textsuperscript{*} \quad
Gang Chen\textsuperscript{*} \quad
Holger Caesar \quad
Javier Alonso\hyp{}Mora\textsuperscript{\dag}\\[2pt]
Delft University of Technology\\
Mekelweg 5, 2628 CD Delft, Netherlands\\
{\tt\smallscriptsize
\{J.Kim-18, G.Dumont, X.Gao-14\}@student.tudelft.nl,
\{G.Chen-5, H.Caesar, J.AlonsoMora\}@tudelft.nl
}%
}}

\begin{document}
\maketitle
\begingroup
\renewcommand\thefootnote{\fnsymbol{footnote}}
\footnotetext[1]{Equal contribution}
\footnotetext[2]{Corresponding Author}
\endgroup

\begin{abstract}
Dense 3D semantic occupancy perception is critical for mobile robots operating in pedestrian-rich environments, yet it remains underexplored compared to its application in autonomous driving. To address this gap, we present MobileOcc, a semantic occupancy dataset for mobile robots operating in crowded human environments. Our dataset is built using an annotation pipeline that incorporates static object occupancy annotations and a novel mesh optimization framework explicitly designed for human occupancy modeling. It reconstructs deformable human geometry from 2D images, then refines and optimizes it using associated LiDAR point data. Using MobileOcc, we establish benchmarks for two tasks: i) Occupancy prediction and ii) Pedestrian velocity prediction, using different methods, including monocular, stereo, and panoptic occupancy, with metrics and baseline implementations for reproducible comparison. Beyond occupancy prediction, we further assess our annotation method on 3D human pose estimation datasets. Results demonstrate that our method exhibits robust performance across different datasets. Our code and dataset are released at \url{https://autonomousrobots.nl/paper_websites/mobileocc}

\end{abstract}    
\section{Introduction}
\label{sec:intro}

\begin{figure}[th!]
  \centering
  \includegraphics[width=\linewidth]{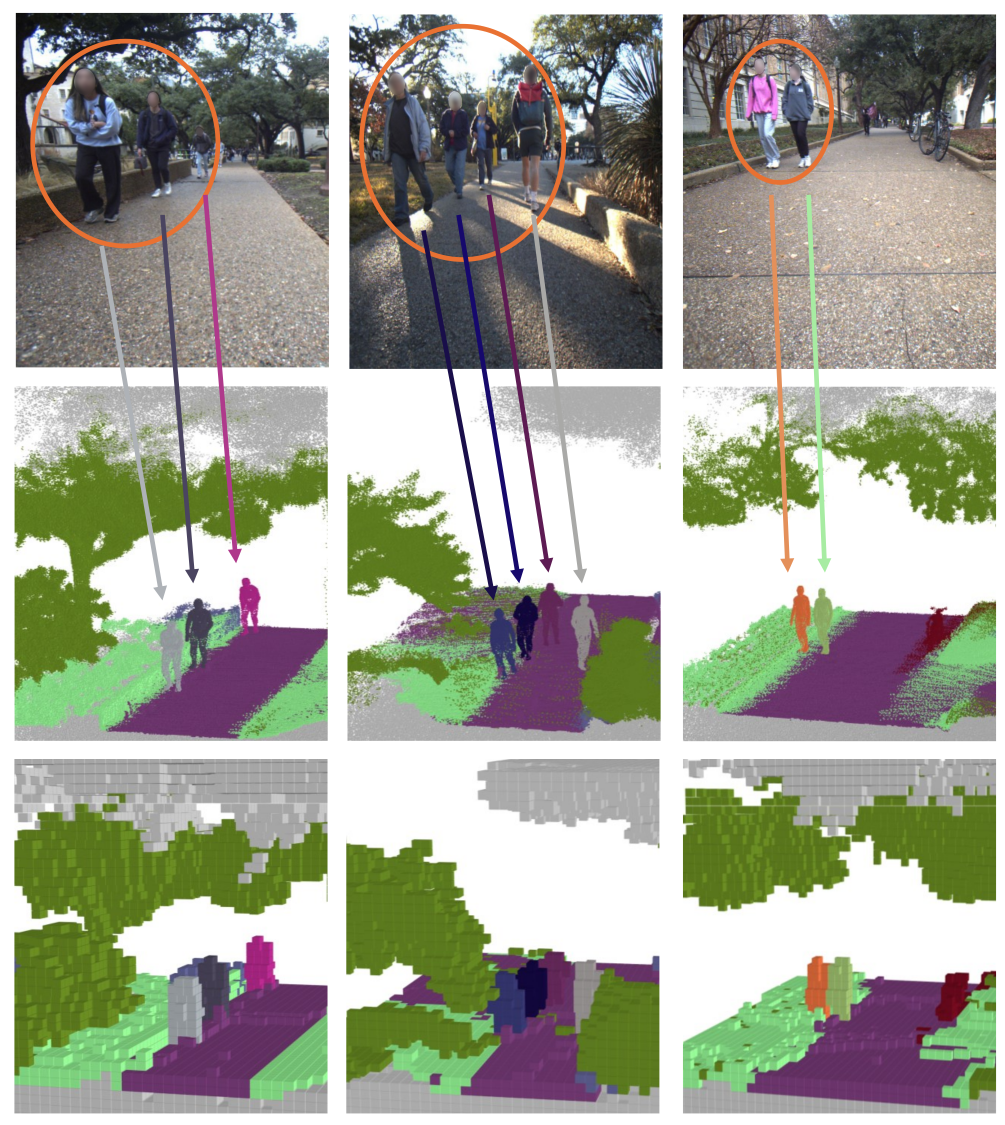}
  \caption{Qualitative results across occupancy resolutions. Top: input image from UT Campus Object Dataset (CODa)~\cite{zhang2024toward}. Middle: our semantic occupancy label at fine resolution (0.02 m). Bottom: semantic occupancy at coarse resolution (0.2 m). Gray voxels represent unknown regions, while free space is not visualized for clarity.
  }  
  \label{fig:first_figure}
\end{figure}

Semantic occupancy prediction~\cite{li2023voxformer,miao2023occdepth,huang2023tri,wei2023surroundocc,tong2023scene} has emerged as a key perception component in modern navigation systems. It predicts a 3D field representing free, occupied, and unknown space, along with semantic labels, directly from image inputs. This enables the construction of a differentiable perception model that can be seamlessly integrated into downstream planning tasks. 
However, existing datasets~\cite{tian2023occ3d, wang2023openoccupancy, li2024sscbench} for training semantic occupancy prediction models are primarily developed for autonomous driving scenarios and focus on rigid objects such as roads, buildings, and vehicles. 
In contrast, non-rigid objects, such as humans, appear less frequently and are typically modeled by aggregating multi-view points into rigid bodies, neglecting their deformable nature.

For mobile robots operating in human-centered environments, this limitation is critical. Humans are among the most frequently encountered entities in navigation, and accurate occupancy prediction must capture their expressive geometry. This is essential for socially aware navigation in dense crowds, interaction-centric navigation among pedestrians, and physical human–robot interaction requiring limb-level, pose-dependent representations.

In this paper, we present MobileOcc, a new semantic occupancy dataset designed for human-populated environments. The dataset is annotated using a proposed pipeline that fuses synchronized image and LiDAR data to generate occupancy labels, with a particular focus on accurate human representation. The pipeline produces occupancy annotations for static objects and for free and unknown space by fusing accumulated and filtered semantic LiDAR points within an OctoMap framework~\cite{hornung2013octomap}.
For human occupancy modeling, we introduce a mesh optimization method that first detects humans in images to estimate their initial pose and mesh, and then optimizes the mesh by fusing corresponding LiDAR points. Compared with image-only human mesh recovery (HMR) methods, which suffer from depth and scale ambiguities, and LiDAR-only approaches, which are limited by sparse points, missed detections, and pose uncertainty in cluttered or distant scenes~\cite{li2022lidarcap, yang2021s3, fan2025lidar}, our fusion-based method produces robust and accurate human meshes by leveraging the strengths of both modalities. 

The contributions of this paper are as follows: 
\begin{itemize}
    \item We introduce MobileOcc, a 3D semantic occupancy prediction dataset built on raw sensor data from UT Campus Object Dataset (CODa)~\cite{zhang2024toward} for mobile robots navigating among near-field pedestrians.
    \item We develop an annotation pipeline that generates occupancy annotations from synchronized RGB images and dense LiDAR data. In particular, the pipeline leverages both image data and LiDAR point clouds to produce accurate representations of human occupancy.
    \item We establish MobileOcc as a benchmark suite by evaluating existing occupancy frameworks on two tasks: i) semantic occupancy prediction and ii) pedestrian velocity prediction, with baselines across monocular, stereo, and panoptic occupancy methods.
\end{itemize}

\section{Related Work}
\label{sec:Lit_review}

\subsection{3D Occupancy Prediction Datasets}

3D occupancy prediction is important for navigation and is widely used in autonomous driving. SemanticKITTI~\cite{behley2019semantickitti} introduced an occupancy perception dataset based on the KITTI dataset, but it does not account for moving objects and has limited scale and diversity. Subsequent datasets such as Occ3D~\cite{tian2023occ3d}, OpenOccupancy~\cite{wang2023openoccupancy}, and SSCBench~\cite{li2024sscbench} provide larger-scale, densely annotated voxel labels on nuScenes~\cite{caesar2020nuscenes}, Waymo~\cite{sun2020scalability}, and KITTI-360~\cite{behley2019semantickitti}. In off-road settings, WildOcc~\cite{zhai2024wildocc} extends RELLIS 3D~\cite{jiang2021rellis} with dense semantic occupancy for unstructured environments. \cref{tab:dataset_comparison} summarizes key differences. These datasets typically model moving vehicles as rigid bodies and accumulate points using pre-labeled 3D bounding-box poses. However, humans are non-rigid and poorly represented in these datasets. For mobile robots, humans are among the most common dynamic objects encountered, and a dataset that accurately models human motion remains needed. Moreover, at fine resolution (\cref{fig:first_figure}), mesh-based labels preserve limb-level geometry that rigid bounding-box approximations cannot capture, which is critical for navigation in tight crowd gaps.

\begin{table}
\centering
\caption{Comparison with existing 3D occupancy datasets. MobileOcc is the first to provide non-rigid human occupancy with per-instance tracking for mobile robot navigation.}
\label{tab:dataset_comparison}
\scriptsize
\setlength{\tabcolsep}{4pt}
\begin{tabular}{l c c c c}
\toprule
Dataset & Domain & Human Model & Instance Track & Non-rigid \\
\midrule
SemanticKITTI~\cite{behley2019semantickitti} & Driving & -- & \ding{55} & \ding{55} \\
Occ3D~\cite{tian2023occ3d} & Driving & Rigid bbox & \ding{55} & \ding{55} \\
OpenOccupancy~\cite{wang2023openoccupancy} & Driving & Rigid bbox & \ding{55} & \ding{55} \\
SSCBench~\cite{li2024sscbench} & Driving & Rigid bbox & \ding{55} & \ding{55} \\
OpenScene~\cite{openscene2023} & Driving & Rigid bbox & \ding{51} & \ding{55} \\
WildOcc~\cite{zhai2024wildocc} & Off-road & -- & \ding{55} & \ding{55} \\
\midrule
MobileOcc (Ours) & Pedestrian & SMPL mesh & \ding{51} & \ding{51} \\
\bottomrule
\end{tabular}
\end{table}

\subsection{3D Occupancy Prediction Methods}

Unlike traditional filtering-based methods that fuse 3D points to generate an occupancy map~\cite{10.1109/TRO.2025.3526084}, occupancy prediction learns to map 2D images to voxel-level semantic occupancy~\cite{tong2023scene}. Camera-only methods such as VoxFormer~\cite{li2023voxformer} and OccDepth~\cite{miao2023occdepth} infer depth cues and decode a dense 3D volume. Surround-view approaches~\cite{huang2023tri,wei2023surroundocc,tong2023scene} aggregate multi-camera features to improve coverage. Efficient variants such as FlashOcc~\cite{yu2023flashocc} and Panoptic-FlashOcc~\cite{yu2024panoptic} keep features in BEV and lift logits to 3D. Our setting is complementary: we focus on mobile robots in human-rich environments and evaluate such frameworks under non-rigid human motion.

\subsection{Image-Based Human Mesh Recovery}
Estimating a 3D human body mesh from a single RGB image has been widely studied using parametric models such as SMPL~\cite{loper2023smpl}. Regression-based methods~\cite{kanazawa2018end,kolotouros2019learning,li2022cliff} directly predict shape and pose parameters, while alternative designs lift skeletons or heatmaps to meshes~\cite{choi2020pose2mesh,moon2020i2l}, or use transformer regressors~\cite{lin2021end,lin2021mesh}. However, purely image-based pipelines suffer from depth ambiguity. To reduce this uncertainty, LiDAR-based methods~\cite{li2022lidarcap,fan2025lidar,zhang2024lidarcapv2} fit SMPL meshes to point clouds, while S3~\cite{yang2021s3} uses implicit fields. LiDAR-only methods can be brittle when human points overlap with nearby objects or become sparse at long distances. We present a training-free method that first estimates an initial mesh from an image, then refines it by fusing LiDAR data, leveraging both modalities and generalizing directly across datasets.

\section{MobileOcc Dataset Building}
\label{sec:method}

Our dataset is constructed by annotating the image and LiDAR data from the CODa dataset~\cite{zhang2024toward} using the proposed annotation pipeline. This section introduces the pipeline by first providing a general overview of the workflow and then detailing its key components.

\begin{figure*}[t!]
  \centering
  \includegraphics[width=\linewidth]{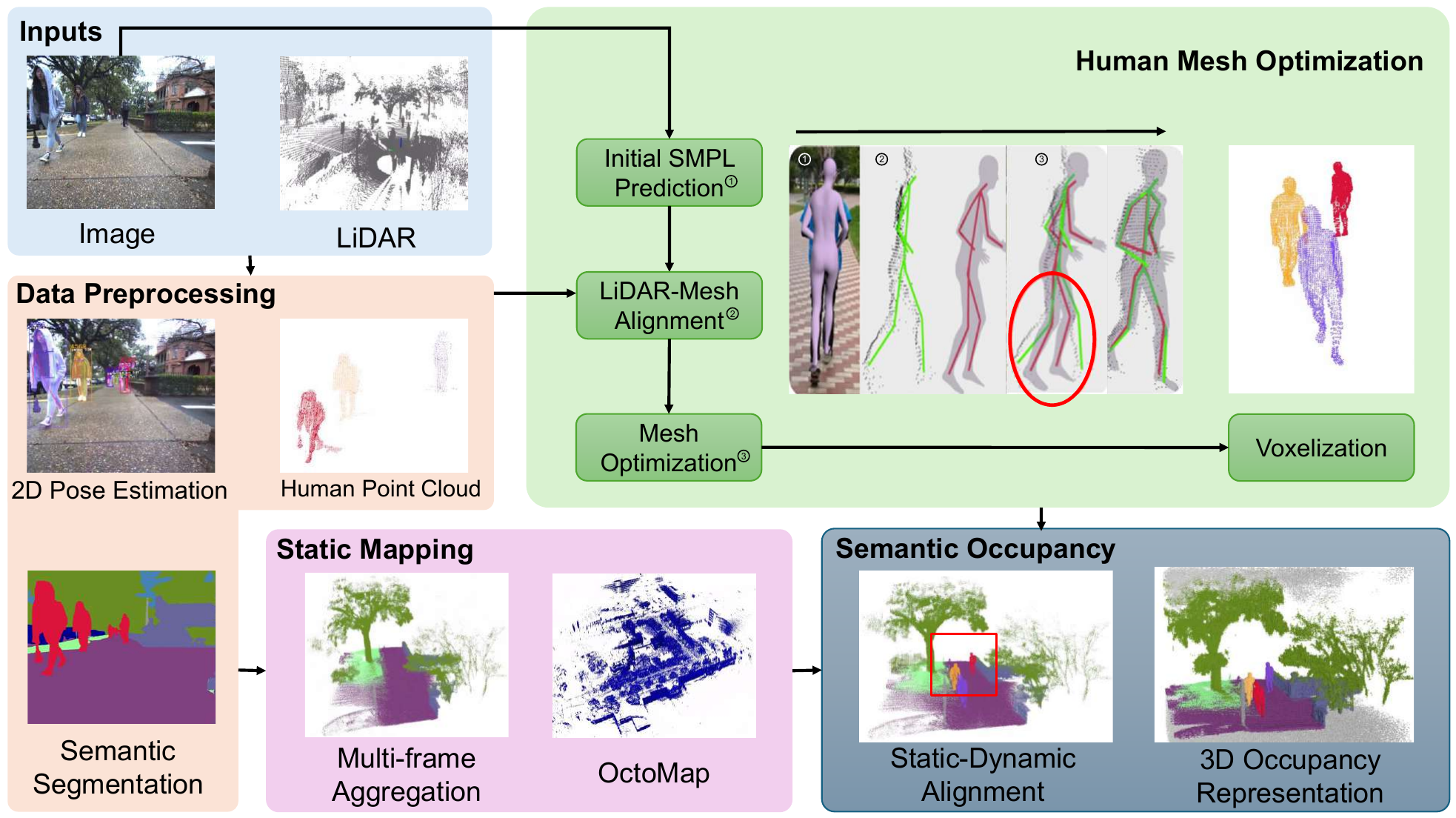}
  \caption{Overview of MobileOcc pipeline. The pipeline consists of: Data preprocessing, Static mapping, Human mesh optimization, and 3D occupancy representation generation. The colors of the voxels follow the labeling scheme used in the Cityscapes dataset~\cite{cordts2016cityscapes}. Human instances are assigned colors, and unknown regions are shown in gray. In human mesh optimization, the green and red skeletons represent the ground truth and the predicted 3D pose of the mesh, respectively.}  
  \label{fig:full_pipeline}
\end{figure*}

\subsection{Annotation Pipeline}
Our general annotation pipeline is shown in~\cref{fig:full_pipeline}.
Given synchronized image streams and LiDAR along a robot trajectory, we first run OC-SORT~\cite{cao2023observation} with YOLOX~\cite{ge2021yolox} detection to obtain tracked 2D pedestrian bounding boxes with IDs. For each tracked person, we estimate 2D pose with keypoint confidence using VITPose~\cite{xu2022vitpose}, obtain instance masks via Mask2Former~\cite{cheng2022masked}, and associate each mask with the LiDAR point cloud. We apply semantic segmentation to assign Cityscapes~\cite{cordts2016cityscapes} class labels for static background elements. All preprocessed data are time-stamped and co-referenced across modalities. With these preprocessing steps, we perform human mesh optimization (\cref{subsec:smpl_dynamic}) and static mapping (\cref{subsec:static}), then fuse them into a 3D occupancy representation (\cref{subsec:occupancy}).

\subsection{Human Mesh Optimization}
\label{subsec:smpl_dynamic}

We propose an optimization-based framework that fuses LiDAR points with monocular images to improve 3D human mesh estimation. Our approach first obtains an initial parametric mesh (SMPL) from the image~\cite{loper2023smpl}, then refines it in three stages: i) filtering out occluded body parts using 2D keypoint cues, ii) performing a coarse mesh alignment to LiDAR via iterative closest point (ICP), and iii) optimizing the SMPL pose/shape parameters to fit the image and LiDAR observations jointly.

\noindent We use the following notation throughout the paper:
\begin{itemize}
  \item \textbf{SMPL parameters:} pose $\boldsymbol{\theta}=\{\boldsymbol{\theta}_{\mathrm{glob}},\,\boldsymbol{\theta}_{\mathrm{body}}\}$, shape $\boldsymbol{\beta}$, and camera translation $\mathbf{t}_{\mathrm{cam}}\!\in\!\mathbb{R}^3$.
  \item \textbf{Camera:} intrinsics $\mathbf{K}$.
  \item \textbf{Mesh:} $\mathcal{M}$ with vertex matrix $\mathbf{M}(\boldsymbol{\beta},\boldsymbol{\theta})\in\mathbb{R}^{6890\times3}$; visible subset $\mathcal{V}\subset\mathcal{M}$ with $n=|\mathcal{V}|$.
  \item \textbf{LiDAR:} point cloud $\mathcal{P}=\{\mathbf{p}_j\}_{j=1}^{m}$ with $m=|\mathcal{P}|$.
  \item \textbf{Operators:} projection $\Pi_{\mathbf{K}}(\cdot)$, global rotation $\mathbf{R}(\boldsymbol{\theta})$, and SMPL joints $\mathbf{J}_i(\boldsymbol{\beta})$.
\end{itemize}

\subsubsection{Initial SMPL Prediction}
We obtain an initial SMPL estimate using CLIFF~\cite{li2022cliff}, a top-down HMR regressor that predicts shape $\boldsymbol{\beta}^0$, pose $\boldsymbol{\theta}^0$, and translation $\mathbf{t}^0$ from a person crop conditioned on its full-image bounding box location. This provides a reasonable initial pose and location in the camera frame, but like all monocular methods, it suffers from depth ambiguity, which our subsequent steps correct.

\subsubsection{LiDAR-Mesh Alignment}
Before optimizing the human mesh provided by CLIFF, we first perform visibility filtering to identify the mesh vertices visible in the LiDAR data.
We then coarsely register the visible mesh vertices $\mathcal{V}$ to the LiDAR points $\mathcal{P}$ via ICP~\cite{besl1992method}, estimating a rigid transform $\mathbf{T}\!\in\!SE(3)$ initialized from $(\boldsymbol{\theta}_{\mathrm{glob}},\mathbf{t}_{\mathrm{cam}})$. Applying $\mathbf{T}$ yields an updated global pose $(\boldsymbol{\theta}_{\mathrm{glob}}',\mathbf{t}_{\mathrm{cam}}')$ that places the mesh correctly in 3D space. However, since ICP is rigid and does not alter internal joint angles, the pose can become inconsistent with the image. Our final optimization step resolves these inconsistencies by adjusting pose and shape using both image and LiDAR cues. 

\subsubsection{Mesh Optimization}

To resolve the remaining pose inconsistencies highlighted in~\cref{fig:full_pipeline} (red ellipse), we refine all SMPL parameters $(\boldsymbol{\beta}, \boldsymbol{\theta}, \mathbf{t}_{\text{cam}})$ to better fit both the 2D and 3D data. Inspired by SMPLify~\cite{bogo2016keep}, we formulate this as minimizing an objective function
\begin{equation}
\label{eq:total}
\begin{aligned}
\mathcal{L}_{\text{total}}(\boldsymbol{\beta}, \boldsymbol{\theta}, \mathbf{t}_{\mathrm{cam}})
= {}& \mathcal{L}_{J}
+ \lambda_{3D}\,\mathcal{L}_{3D}
+ \lambda_{\theta}\,\mathcal{L}_{\theta}
+ \lambda_{a}\,\mathcal{L}_{a}
\\
&{}+ \lambda_{\beta}\,\mathcal{L}_{\beta}
+ \lambda_{\mathrm{occ}}\,\mathcal{L}_{\theta,\mathrm{occ}}
\end{aligned}
\end{equation}
that contains terms for image alignment, 3D alignment, and regularization priors with scalar weights $\lambda$:

\begin{itemize}
    \item \textbf{2D joint reprojection} ($\mathcal{L}_{J}$): Penalizes the distance between the projected 3D joints of the mesh and the detected 2D keypoints in the image,
    \begin{equation}
    \label{eq:reproj}
    \mathcal{L}_{J}=\sum_{i} w_i\, \rho\!\Big(\Pi_{\mathbf{K}}\big(\mathbf{R}(\boldsymbol{\theta})\,\mathbf{J}_i(\boldsymbol{\beta})\big) + \mathbf{t}_{\mathrm{cam}} - \mathbf{J}^{\mathrm{est}}_{i}\Big),
    \end{equation}
    where $w_i$ is the keypoint confidence and $\rho(\cdot)$ is the Geman–McClure robust penalty~\cite{geman1986bayesian}. 
    
    \item \textbf{3D LiDAR alignment} ($\mathcal{L}_{3D}$): Minimizes the distance between visible mesh vertices $\mathcal{V}$ and the LiDAR point cloud $\mathcal{P}$ using a symmetric, size-normalized Chamfer distance:
    \begin{equation}
    \label{eq:chamfer}
    \mathcal{L}_{3D} =
    \frac{1}{|\mathcal{V}|}\!\sum_{\mathbf{v}\in\mathcal{V}} \min_{\mathbf{p}\in\mathcal{P}} \|\mathbf{v}-\mathbf{p}\|_2^2
    \;+\;
    \frac{1}{|\mathcal{P}|}\!\sum_{\mathbf{p}\in\mathcal{P}} \min_{\mathbf{v}\in\mathcal{V}} \|\mathbf{p}-\mathbf{v}\|_2^2.
    \end{equation}
    Optimizing over $\boldsymbol{\theta}$ and $\boldsymbol{\beta}$ lets this term correct local pose/shape errors that pure 2D fitting cannot.
    
    \item \textbf{Pose prior} ($\mathcal{L}_{\theta}$): The SMPLify MoG pose prior~\cite{bogo2016keep} discourages implausible joint configurations,
    \begin{equation}
    \mathcal{L}_{\theta} = -\log \sum_{j=1}^{N} g_j \,\mathcal{N}(\boldsymbol{\theta};\,\boldsymbol{\mu}_{\theta,j},\mathbf{\Sigma}_{\theta,j}).
    \end{equation}
    
    \item \textbf{Anti-hyperextension prior} ($\mathcal{L}_{a}$): Discourages unnatural positive bending at elbows/knees (as in~\cite{bogo2016keep}),
    \begin{equation}
    \mathcal{L}_{a} = \sum_{k\in\{\text{knees, elbows}\}} \exp(\theta_k).
    \end{equation}
    
    \item \textbf{Shape prior} ($\mathcal{L}_{\beta}$): A quadratic prior on shape parameters keeps $\boldsymbol{\beta}$ within the SMPL space~\cite{loper2023smpl},
    \begin{equation}
    \mathcal{L}_{\beta}=\boldsymbol{\beta}^{\top}\mathbf{\Sigma}_{\beta}^{-1}\boldsymbol{\beta}.
    \end{equation}
    
    \item \textbf{Occlusion-aware pose consistency prior} ($\mathcal{L}_{\theta,\mathrm{occ}}$): For joints flagged invisible by the visibility filter, we regularize deviation from the initialization to prevent unrealistic drift. Let $\mathcal{I}$ be the set of occluded joints with initial and current unit quaternions $\{\mathbf{q}_i^{(0)},\mathbf{q}_i\}$; we use
    \begin{equation}
    \label{eq:occ}
    \mathcal{L}_{\theta,\mathrm{occ}} = \sum_{i\in\mathcal{I}} \Big(1 - \langle \mathbf{q}_i^{(0)}, \mathbf{q}_i \rangle^2\Big).
    \end{equation}
\end{itemize}

We optimize $\mathcal{L}_{\text{total}}$ with respect to $(\boldsymbol{\beta},\boldsymbol{\theta},\mathbf{t}_{\mathrm{cam}})$ using gradient descent in PyTorch. The result is a refined set $(\hat{\boldsymbol{\beta}}, \hat{\boldsymbol{\theta}}, \mathbf{t}_{\mathrm{cam}}^{*})$ that balances fitting the image (2D joints) and fitting the LiDAR (3D points), while respecting human shape and pose priors. In contrast to SMPLify~\cite{bogo2016keep}, we omit the interpenetration term and instead add $\mathcal{L}_{3D}$ (LiDAR-mesh Chamfer) and $\mathcal{L}_{\theta,\mathrm{occ}}$ (occlusion-aware pose consistency), while keeping $\mathcal{L}_{J}$, $\mathcal{L}_{\theta}$, $\mathcal{L}_{a}$, $\mathcal{L}_{\beta}$. We drop interpenetration due to its high cost and limited impact on final accuracy~\cite{kolotouros2019learning}. Our occlusion-aware visibility and pose-consistency terms help keep occluded limbs in reasonable positions without explicit handling of interpenetration. Moreover, since we voxelize occupancy from the final mesh surface (i.e., the exterior boundary), interpenetration primarily results in rare internal self-overlaps and has minimal effect on the occupied/free voxel boundary.

\subsection{Static Map Generation}
\label{subsec:static}

We build a high-resolution static semantic map by fusing multi-frame LiDAR geometry with camera-derived semantics, while explicitly filtering dynamic pedestrians, including those outside the camera frustum.

\subsubsection{LiDAR-based Detection}

To avoid moving pedestrians being imprinted into the background, we run a LiDAR 3D detector per sweep and use its pedestrian outputs, together with the image-based pedestrian detection result, as a mask during static mapping. 
Points that fall within these masks are excluded from static map generation. This stage neither changes semantic labels nor trains any model, and its sole purpose is to suppress pedestrian evidence before building the static map.

\subsubsection{Multi-frame Aggregation}
We fuse LiDAR geometry and camera semantics over time in a global frame. Each sweep is pose-compensated, and image semantics are lifted to the corresponding 3D points, which are then inserted into a voxel map that maintains compact per-voxel label counts. As frames accumulate, each voxel label is updated through max-voting, and geometric sanity checks (e.g., ground consistency) reject outliers. After processing the sequence, the final static map is obtained by assigning a semantic label to each voxel, yielding a human-free background that we later combine with human meshes to obtain human-aware occupancy.

\subsection{3D Occupancy Representation}
\label{subsec:occupancy}

\subsubsection{Static-dynamic Alignment}

We merge the human-free static map (\cref{subsec:static}) with pedestrian evidence (\cref{subsec:smpl_dynamic}) in a shared robot-local frame as shown in~\cref{fig:full_pipeline}. For each frame, we transform the current static points and rasterized meshes into a common local coordinate frame using the synchronized robot pose, then voxelize them. Static voxels inherit Cityscapes~\cite{cordts2016cityscapes} semantics from RGB points. Pedestrian voxels carry per-instance identifiers, as we assign each tracked pedestrian a unique ID label. 

\subsubsection{Label Assignment}
After alignment, we instantiate voxel grids in the robot frame with a user-specified resolution (up to 0.02 m) and assign labels in a priority order: dynamic humans take precedence over static background, so pedestrians do not get imprinted into the background during fusion. If no pedestrian points are present, we assign static semantics using a majority-label-wins rule to avoid label oscillation when many points land in the same voxel. Remaining cells are completed using an OctoMap query~\cite{hornung2013octomap}, which contributes free/unknown states so that occupied space is supported by explicit geometry. This produces an aligned voxel grid in which static semantics and dynamic human occupancy coexist, and all remaining space is consistently categorized as either free or unknown.

We output our final data in the NuScenes~\cite{caesar2020nuscenes} dataset format to ensure easier integration and consistency with existing tools.
\section{Results}
\label{sec:Experiment}

In this section, we present a comprehensive evaluation of our approach. We first evaluate human mesh optimization on three datasets. We then validate the quality of the generated occupancy labels against available ground-truth annotations and quantify the robustness of upstream detection. Finally, we train representative baselines on MobileOcc and report performance on two tasks: i) occupancy prediction and ii) pedestrian velocity prediction.

\subsection{Human Mesh Evaluation}
We evaluate our human mesh optimization method in 3DPW~\cite{von2018recovering}, SLOPER4D~\cite{dai2023sloper4d}, and HumanM3~\cite{fan2023human} datasets. We report four standard metrics: PVE (Per-Vertex Error, mm), the average vertex-level Euclidean distance; MPJPE (Mean Per-Joint Position Error, mm) over the dataset's kinematic joint set; PA-MPJPE (Procrustes-Aligned MPJPE, mm), which isolates articulated pose quality from global placement; and MPERE~\cite{fan2025lidar} (Mean Per-Edge Relative Error), the average relative error of mesh edge lengths. PVE and MPERE are reported only when the vertex ground truth is available.

\subsubsection{Results on the 3DPW Dataset}

\label{subsec:sim_3dpw}
3DPW~\cite{von2018recovering} dataset is a popular in-the-wild dataset with 3D pose and shape ground truth. Since no LiDAR points are available, we add synthetic LiDAR sweeps at different densities in the camera frame. Specifically, we mimic Ouster-style sensors with vertical beams = {32, 64, 128} (higher $\rightarrow$ denser body coverage) and added Gaussian range perturbations and probabilistic point dropouts following the CARLA LiDAR noise model~\cite{dosovitskiy2017carla}.
This allows us to i) isolate the effect of adding LiDAR points to image/video HMR; and ii) study how performance scales with sensor density. 

\cref{tab:3dpw-sim} presents the results of our method compared with other image- and video-based HMR approaches evaluated on the 3DPW dataset. Since image- and video-based methods inherently lack depth, they are evaluated using root alignment~\cite{von2018recovering}, which translates the predicted pelvis to the ground-truth position. In contrast, our method instead uses simulated LiDAR on the human body to optimize position, pose, and shape directly in the sensor frame.

The results show that our method improves in all metrics without using root alignment, indicating better absolute placement (PVE/MPJPE) and pose accuracy (PA-MPJPE). Performance increases with denser LiDAR, which provides more reliable 3D correspondences.

\begin{table}[t!]
\centering
\caption{Simulated LiDAR comparison on 3DPW~\cite{von2018recovering}. Video/Image methods use no depth and are reported with root alignment. Ours fuses simulated LiDAR without root alignment.}
\scriptsize
\setlength{\tabcolsep}{3pt}
\begin{tabular}{@{}ll
                S[table-format=3.1]
                S[table-format=3.1]
                S[table-format=3.1]@{}}
\toprule
\textbf{Modality} & \textbf{Method} & {\textbf{PVE}\,$\downarrow$} & {\textbf{MPJPE}\,$\downarrow$} & {\textbf{PA-MPJPE}\,$\downarrow$} \\
\midrule
\multirow{4}{*}{Video}
  & VIBE~\cite{kocabas2020vibe}            & 99.1 & 82.9 & 51.9 \\
  & DynaBOA~\cite{guan2022out}             & 82.0 & 65.5 & 40.4 \\
  & MotionBERT~\cite{zhu2023motionbert}    & 79.4 & 68.8 & 40.6 \\
  & WHAM\mbox{-}B~\cite{shin2024wham}      & 71.0 & 59.4 & 37.2 \\
\addlinespace[1pt]
\multirow{8}{*}{Image}
  & SMPLify~\cite{bogo2016keep}            & 106.8 & \multicolumn{1}{c}{--} & \multicolumn{1}{c}{--} \\
  & TRACE~\cite{sun2023trace}              & 97.3  & 79.1 & 37.8 \\
  & SPIN~\cite{kolotouros2019learning}     & 96.9  & 59.2 & \multicolumn{1}{c}{--} \\
  & ROMP~\cite{sun2021monocular}           & 93.4  & 76.7 & 47.3 \\
  & HybrIK~\cite{li2021hybrik}            & 82.3  & 71.6 & 41.8 \\
  & CLIFF~\cite{li2022cliff}              & 81.2  & 69.0 & 43.0 \\
  & Cha.~\cite{cha2022multi}              & 76.3  & 66.0 & 39.0 \\
  & PLIKS~\cite{shetty2023pliks}          & 73.3  & 60.5 & 38.5 \\
\addlinespace[1pt]
\multirow{3}{*}{LiDAR}
  & Ours (Ouster\mbox{-}32)               & 73.2  & 57.0 & 47.7 \\
  & Ours (Ouster\mbox{-}64)               & 57.2  & 43.9 & 38.5 \\
  & \textbf{Ours (Ouster\mbox{-}128)}     & \bfseries 50.5 & \bfseries 39.1 & \bfseries 35.1 \\
\bottomrule
\end{tabular}
\label{tab:3dpw-sim}
\end{table}

\subsubsection{Results on the SLOPER4D and HumanM3 Datasets}
\label{subsec:real_lidar}

We further evaluate our method on SLOPER4D~\cite{dai2023sloper4d} and HumanM3~\cite{fan2023human} datasets, which are targeted for LiDAR-based 3D human pose estimation. SLOPER4D~\cite{dai2023sloper4d} has synchronized RGB image and LiDAR points, and ground-truth mesh vertices. An Ouster-OS1-128 LiDAR is used for data collection, and only one person appears in the dataset at a time. HumanM3~\cite{fan2023human} has multiple humans at a distance, and a Livox Mid-100 LiDAR is used. It has ground-truth for 3D joints but not for the mesh.

On SLOPER4D, we report PVE, MPJPE, PA-MPJPE, and MPERE~\cite{fan2025lidar}. MPERE is computed for any method that outputs a mesh (SMPL-based or model-free that reconstructs surfaces). On HumanM3, vertex ground truth is unavailable, so we report only joint metrics (MPJPE and PA-MPJPE). At test time, we reuse the 3DPW pipeline: visibility filtering, single-sweep rigid pre-alignment, and joint refinement using RGB and LiDAR, all evaluated in the sensor frame.

We compare against LiDAR methods benchmarked on these two datasets and present additional results using our base image model, CLIFF~\cite{li2022cliff}. As summarized in~\cref{tab:real}, our method achieves strong accuracy without any LiDAR-specific training. The PA-MPJPE columns directly isolate the effect of our LiDAR-guided optimization: comparing CLIFF (RGB) to Ours (RGB+LiDAR) shows substantial reductions on both SLOPER4D and HumanM3, confirming that most gains arise from test-time optimization with LiDAR constraints. Most competing methods are trained on LiDAR (and often dataset-specific) supervision, whereas our improvements arise purely from the optimization process at inference.

\begin{table}[t]
\centering
\scriptsize
\setlength{\tabcolsep}{0.7pt}
\caption{Comparison on real LiDAR: SLOPER4D and HumanM3. Modality: RGB = image, LiDAR = point clouds, RGB+LiDAR = fused. The PA-MPJPE (mm) columns isolate the effect of our LiDAR optimization over the CLIFF initialization.}
\label{tab:real}
\begin{tabular}{@{}l
                S[table-format=3.1]
                S[table-format=3.1]
                S[table-format=2.1]
                S[table-format=1.3]
                S[table-format=3.1]
                S[table-format=2.1]@{}}
\toprule
& \multicolumn{4}{c}{\textbf{SLOPER4D}~\cite{dai2023sloper4d}} & \multicolumn{2}{c}{\textbf{HumanM3}~\cite{fan2023human}} \\
\cmidrule(lr){2-5} \cmidrule(l){6-7}
\textbf{Method} & {\textbf{PVE}\,$\downarrow$} & {\textbf{MPJPE}\,$\downarrow$} & {\textbf{PA-MPJPE}\,$\downarrow$} & {\textbf{MPERE}\,$\downarrow$} & {\textbf{MPJPE}\,$\downarrow$} & {\textbf{PA-MPJPE}\,$\downarrow$} \\
\midrule
\multicolumn{7}{@{}l}{\emph{LiDAR-based}} \\
PRN~\cite{fan2025lidar}              & \multicolumn{1}{c}{--} & 57.0 & \multicolumn{1}{c}{--} & \multicolumn{1}{c}{--} & 82.2 & \multicolumn{1}{c}{--} \\
V2V\mbox{-}PoseNet~\cite{moon2018v2v} & \multicolumn{1}{c}{--} & \bfseries 50.7 & \multicolumn{1}{c}{--} & \multicolumn{1}{c}{--} & 83.0 & \multicolumn{1}{c}{--} \\
LiDAR\mbox{-}HMR~\cite{fan2025lidar} & \bfseries 51.9 & 51.0 & \multicolumn{1}{c}{--} & 0.094 & \bfseries 77.6 & \multicolumn{1}{c}{--} \\
LiDARCap~\cite{li2022lidarcap}       & 148.1 & 158.3 & \multicolumn{1}{c}{--} & \bfseries 0.050 & 175.8 & \multicolumn{1}{c}{--} \\
SAHSR~\cite{jiang2019skeleton}       & 81.2  & 72.6  & \multicolumn{1}{c}{--} & 0.085 & 105.5 & \multicolumn{1}{c}{--} \\
VoteHMR~\cite{liu2021votehmr}        & 60.9  & 54.6  & \multicolumn{1}{c}{--} & 0.079 & 105.8 & \multicolumn{1}{c}{--} \\
\midrule
\multicolumn{7}{@{}l}{\emph{RGB-based}} \\
CLIFF~\cite{li2022cliff}             & 93.7  & 84.7  & 69.3 & 0.057 & 106.3 & 66.5 \\
\midrule
\multicolumn{7}{@{}l}{\emph{RGB+LiDAR}} \\
\textbf{Ours}                        & 64.4  & 55.8  & \bfseries 43.5 & 0.060 & 83.9 & \bfseries 58.1 \\
\bottomrule
\end{tabular}
\end{table}

\newcommand{\citylegend}{
  \small
  \setlength{\fboxsep}{0pt}
  \centering
  \cityclsicon{pedestrian}\, pedestrian \hspace{0.8em}
  \cityclsicon{car}\, car \hspace{0.8em}
  \cityclsicon{otherstruct}\, other struct. \hspace{0.8em}
  \cityclsicon{pole}\, pole \hspace{0.8em}
  \cityclsicon{road}\, road \hspace{0.8em}
  \cityclsicon{terrain}\, terrain \hspace{0.8em}
  \cityclsicon{truck}\, truck \hspace{0.8em}
  \cityclsicon{twowheeler}\, two-wheeler \hspace{0.8em}
  \cityclsicon{vegetation}\, vegetation
}
\newcommand{\occtextwidth}{\textwidth}
\setlength{\dblfloatsep}{24pt plus 2pt minus 2pt}

\begin{figure*}[!t]
  \centering

  \begin{minipage}{\textwidth}
    \centering
    \citylegend
  \end{minipage}

  \vspace{6mm}

  \includegraphics[width=\occtextwidth]{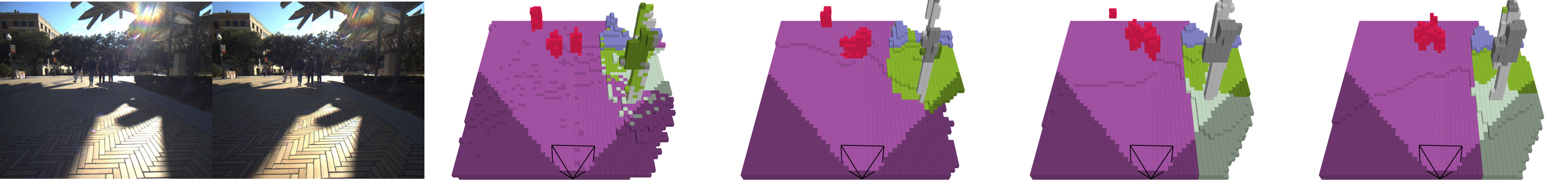}
  \includegraphics[width=\occtextwidth]{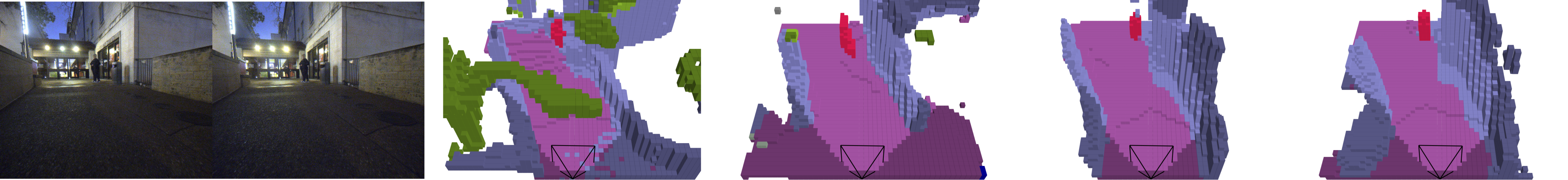}

\noindent
\begin{minipage}{0.273\textwidth} 
    \centering
    \scriptsize\bfseries Stereo (L+R)
\end{minipage}%
\begin{minipage}{0.182\textwidth} 
    \centering
    \scriptsize\bfseries Ground Truth
\end{minipage}%
\begin{minipage}{0.182\textwidth}
    \centering
    \scriptsize\bfseries VoxFormer-T\textnormal{~\cite{li2023voxformer}}
\end{minipage}%
\begin{minipage}{0.182\textwidth}
    \centering
    \scriptsize\bfseries FlashOcc (8f)\textnormal{~\cite{yu2023flashocc}}
\end{minipage}%
\begin{minipage}{0.182\textwidth} 
    \centering
    \scriptsize\bfseries Panoptic-FlashOcc (8f)\textnormal{~\cite{yu2024panoptic}}
\end{minipage}

  \vspace{-1mm}
  \caption{Qualitative comparison of different baselines under various lighting conditions, including sunny, night, and cloudy scenes.}
  \label{fig:occ_comparison}
\end{figure*}

\begin{figure*}[!t]
  \centering

  \includegraphics[width=\occtextwidth]{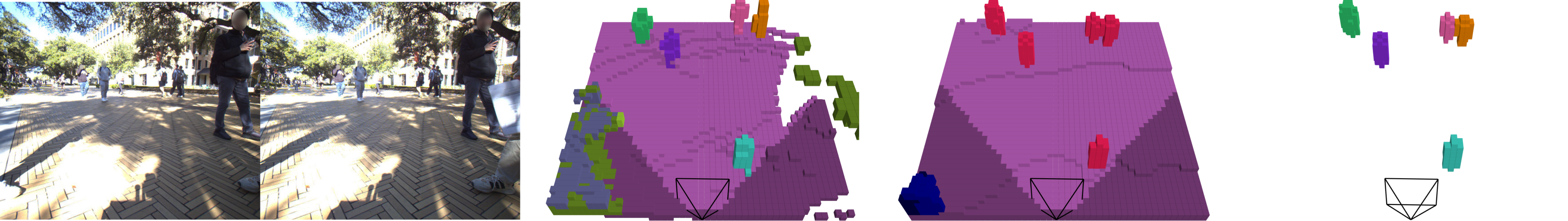}

  \includegraphics[width=\occtextwidth]{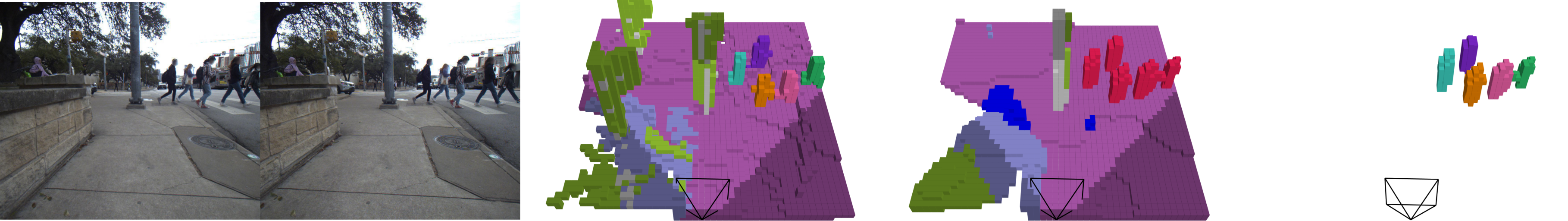}

  \includegraphics[width=\occtextwidth]{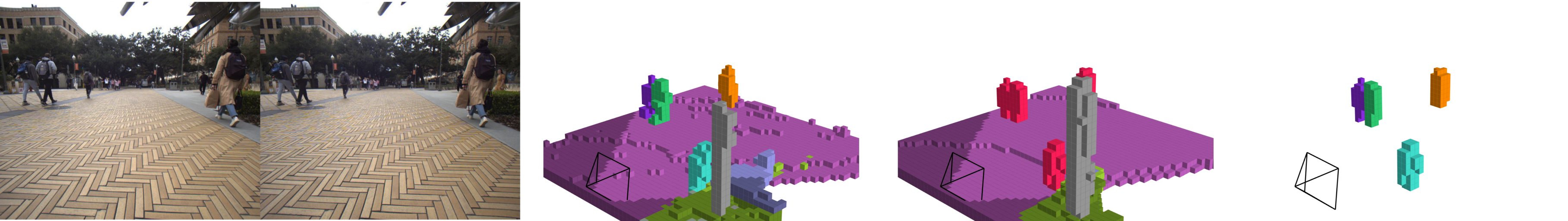}

  \noindent
  \begin{minipage}{0.333\textwidth} 
    \centering
    \scriptsize\bfseries Stereo (L+R)
  \end{minipage}%
  \begin{minipage}{0.222\textwidth} 
    \centering
    \scriptsize\bfseries Ground Truth
  \end{minipage}%
  \begin{minipage}{0.222\textwidth}
    \centering
    \scriptsize\bfseries Predicted Semantic Occupancy
  \end{minipage}%
  \begin{minipage}{0.222\textwidth}
    \centering
    \scriptsize\bfseries Predicted Instances
  \end{minipage}

  \vspace{-1mm}
  \caption{Panoptic occupancy prediction performance using Panoptic-FlashOcc (8f)~\cite{yu2024panoptic}.}
  \label{fig:pano_comparison}

\end{figure*}

\begin{figure*}[!t]
  \centering

  \includegraphics[width=0.49\textwidth]{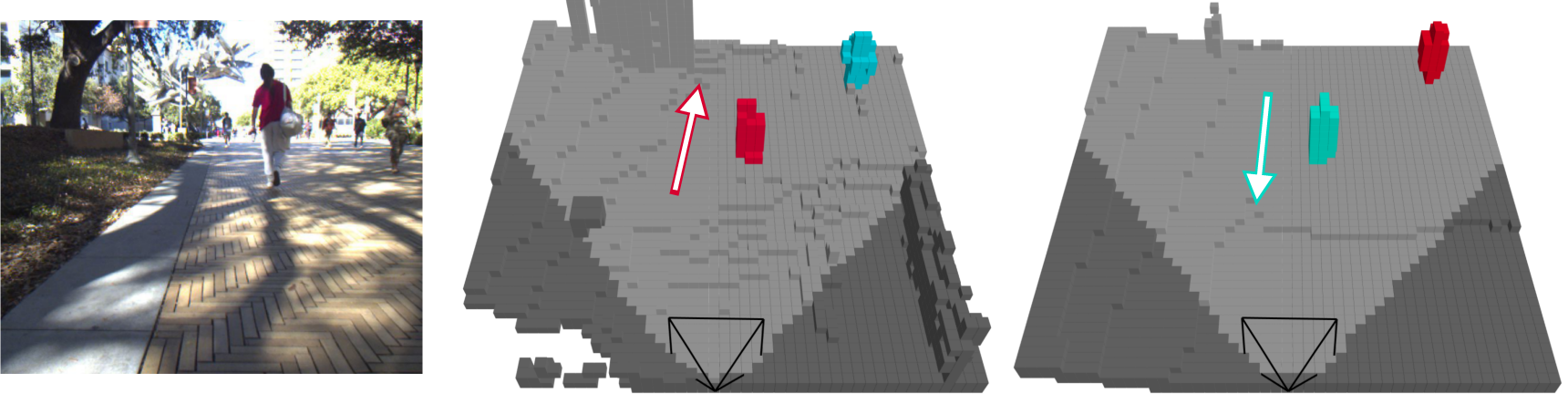}%
  \hfill
  \includegraphics[width=0.49\textwidth]{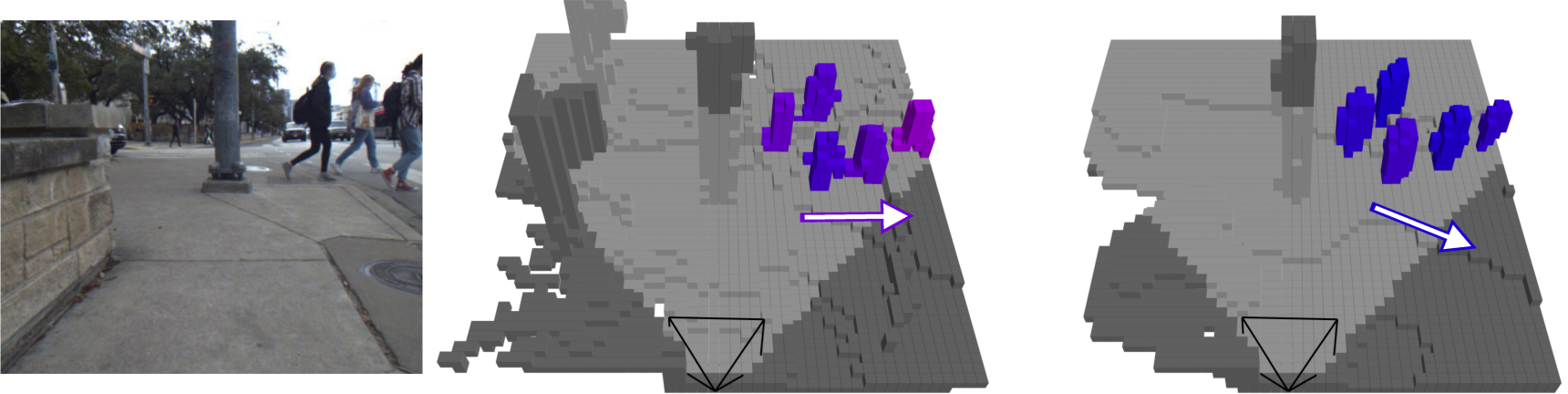}


  \includegraphics[width=0.49\textwidth]{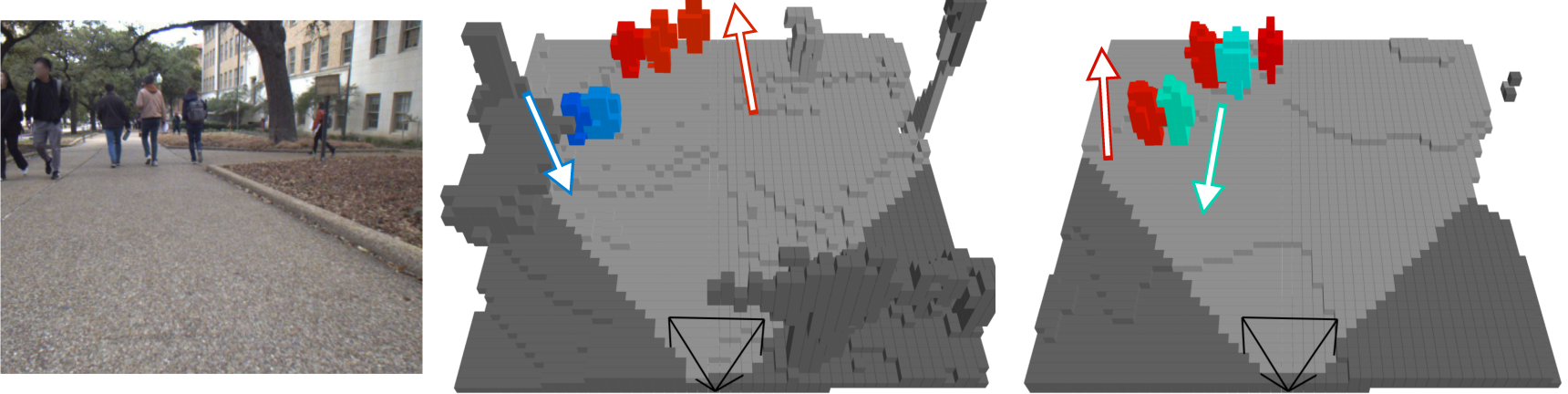}%
  \hfill
  \includegraphics[width=0.49\textwidth]{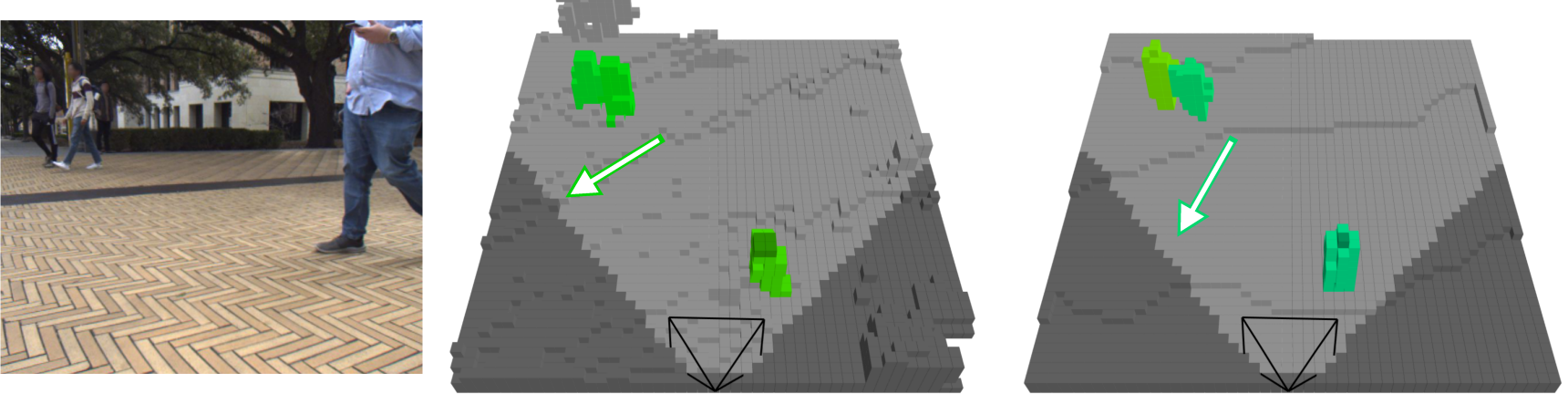}

  \noindent
  \begin{minipage}{0.136\textwidth}
    \centering
    \scriptsize\bfseries Stereo Left
  \end{minipage}%
  \begin{minipage}{0.182\textwidth}
    \centering
    \scriptsize\bfseries Ground Truth
  \end{minipage}%
  \begin{minipage}{0.182\textwidth}
    \centering
    \scriptsize\bfseries Velocity Prediction
  \end{minipage}%
  \begin{minipage}{0.136\textwidth}
    \centering
    \scriptsize\bfseries Stereo Left
  \end{minipage}%
  \begin{minipage}{0.182\textwidth}
    \centering
    \scriptsize\bfseries Ground Truth
  \end{minipage}%
  \begin{minipage}{0.182\textwidth}
    \centering
    \scriptsize\bfseries Velocity Prediction
  \end{minipage}

  \vspace{-1mm}
  \caption{Pedestrian velocity prediction using Panoptic-FlashOcc-vel (8f). The directions are shown in colors and arrows.}
  \label{fig:flow_comparison}
  \vspace{-2mm}
\end{figure*}

\subsection{Occupancy Label Quality}
Pointwise manual semantic annotation of 3D LiDAR data is prohibitively expensive, and CODa provides such labels for only a small portion of its sequences. We validate our voxel labels against CODa ground truth on 4 sequences (400 frames) and assess detection reliability across conditions; quantitative results are reported in \cref{tab:label_quality}.

\subsubsection{Static voxel-label quality}
We evaluate final static semantic voxel labels against CODa's ground-truth semantic LiDAR point annotations for road, vegetation, and terrain. High GT coverage confirms that our voxel grid captures nearly all annotated geometry. The primary error source is taxonomy and domain mismatch between our Cityscapes-trained segmentation and CODa's campus-outdoor categories; e.g., CODa's ``Short Vegetation'' maps to Cityscapes' \textit{terrain} or \textit{vegetation} depending on context, which mainly limits vegetation and terrain IoU.

\subsubsection{Pedestrian occupancy quality}
We evaluate dynamic pedestrian voxel labels using CODa 3D pedestrian bounding boxes as proxy ground truth. Since CODa does not provide dense human-geometry annotations, bounding boxes serve as the best available reference; however, they are inherently conservative because they overcount occupied volume by including empty space around the body and by including ground-plane voxels at pedestrians' feet. We therefore consider only camera-visible occupied voxels within each bbox volume and exclude the bottom 0.1\,m of each bbox, which contains road/terrain voxels rather than body geometry. As shown in~\cref{tab:label_quality}, precision is 84.4\% and recall is 81.9\%, confirming that our labels capture the majority of pedestrian geometry.

\subsubsection{Pedestrian occupancy quality vs.\ LiDAR body GT.}
Bounding boxes overcount occupied volume, so we additionally validate the SMPL-derived voxels against a tighter reference built from raw LiDAR returns. For each pedestrian within the benchmark range, we extract the LiDAR points that fall inside CODa's human-annotated 3D bounding box and voxelize them on the $0.2$~m evaluation grid to obtain a per-pedestrian LiDAR body ground truth, a
sampling of the sensor-facing body surface, restricted to the camera frustum to ensure a like-for-like comparison with our SMPL voxels. LiDAR returns lie on the visible body surface, whereas SMPL voxelization fills the body volume, so a LiDAR voxel naturally sits roughly half a body-depth ($\sim 0.19$~m, measured) ahead of the corresponding SMPL voxel at the body center. We therefore evaluate at the standard $\pm 1$ voxel-size center-to-center tolerance, with precision (SMPL$\to$GT) and recall (GT$\to$SMPL). Tab.~\ref{tab:pedestrian-voxel-quality} reports per-sequence results across $400$ frames and $790$ pedestrian instances. Precision reaches $95.5$ and F1 $92.5$, substantially above the conservative bounding-box proxy, confirming that the pedestrian voxels closely match true body geometry.

\subsubsection{Detection robustness}
To quantify how upstream detection errors propagate into occupancy labels, we perform manual quality assurance on 300 frames spanning crowded, uncrowded, and nighttime conditions (\cref{tab:label_quality}). Daytime detection achieves high recall and precision across both crowd densities. Nighttime recall decreases due to low illumination, while precision remains high, indicating that missed detections dominate false positives. These misses affect only the frames in which they occur and do not propagate into the static map, since pedestrian filtering already excludes detected humans from static fusion and human occupancy overrides static labels during alignment. We distinguish between daytime and nighttime sequences in the dataset, enabling users to select subsets that match their target deployment conditions.

\begin{table}[t]
\centering
\scriptsize
\setlength{\tabcolsep}{2.5pt}
\caption{Occupancy label quality and upstream detection robustness. Top: voxel labels evaluated against CODa ground truth (4 sequences, 400 frames). Bottom: manual detection QA on 300 frames ($T$: correct; $M_\text{vis}$/$M_\text{occ}$: missed visible/occluded(counts); $W$: false positives).}
\label{tab:label_quality}
\begin{tabular}{@{}l*{7}{c}@{}}
\toprule
\multicolumn{8}{@{}l}{\textbf{Static Semantic Voxels} \textit{(vs.\ GT LiDAR points)}} \\[2pt]
\multicolumn{6}{@{}l}{\quad mIoU (road / veg.\ / terrain)} & \multicolumn{2}{r@{}}{71.2 (88.8 / 68.0 / 57.0)} \\
\multicolumn{6}{@{}l}{\quad GT point coverage} & \multicolumn{2}{r@{}}{96.8} \\
\multicolumn{6}{@{}l}{\quad Voxel-level accuracy} & \multicolumn{2}{r@{}}{88.1} \\
\multicolumn{6}{@{}l}{\quad Point-level accuracy} & \multicolumn{2}{r@{}}{94.0} \\[2pt]
\midrule
\multicolumn{8}{@{}l}{\textbf{Pedestrian Voxels} \textit{(vs.\ GT 3D bounding boxes)}} \\[2pt]
\multicolumn{6}{@{}l}{\quad Precision} & \multicolumn{2}{r@{}}{84.4} \\
\multicolumn{6}{@{}l}{\quad Recall} & \multicolumn{2}{r@{}}{81.9} \\
\multicolumn{6}{@{}l}{\quad F1} & \multicolumn{2}{r@{}}{83.1} \\[2pt]
\midrule\midrule
\multicolumn{8}{@{}l}{\textbf{Detection Robustness} \textit{(manual QA, 300 frames)}} \\[2pt]
\quad Condition & $T$ & $M_\text{vis}$ & $M_\text{occ}$ & $W$ & Recall & Precision & Miss$_\text{vis}$\% \\[1pt]
\quad Crowded   & 480 & 6  & 6 & 5 & 98.8 & 99.0 & 1.2 \\
\quad Uncrowded & 191 & 1  & 2 & 0 & 99.5 & 100  & 0.5 \\
\quad Night     &  40 & 12 & 3 & 1 & 76.9 & 97.6 & 23.1 \\
\bottomrule
\end{tabular}
\end{table}

\begin{table}[t]
\centering
\setlength{\tabcolsep}{5pt}
\caption{Pedestrian voxel quality vs.\ LiDAR body GT, per CODa sequence ($400$ frames, $790$ pedestrian instances).}
\renewcommand{\arraystretch}{0.9}
\small
\begin{tabular}{lccccc}
\toprule
Seq & 7 & 11 & 16 & 18 & All \\
\midrule
Precision $\uparrow$ & $95.2$ & $98.9$ & $91.1$ & $96.9$ & $\mathbf{95.5}$ \\
Recall $\uparrow$    & $83.6$ & $87.8$ & $95.6$ & $92.6$ & $\mathbf{89.9}$ \\
F1 $\uparrow$        & $89.0$ & $93.0$ & $93.3$ & $94.7$ & $\mathbf{92.5}$ \\
\bottomrule
\end{tabular}
\label{tab:pedestrian-voxel-quality}
\end{table}

\section{Benchmark}
\subsection{Benchmark Details}
We evaluate all methods at 0.2\,m voxel resolution over a grid spanning $x \in [0.4, 10.0]$\,m, $y \in [-4.8, 4.8]$\,m, $z \in [-1.0, 3.8]$\,m, downsampled to 5\,Hz. The training/validation split contains 92{,}303/24{,}208 frames (79/21\%), of which 30{,}457/7{,}165 have pedestrian annotations. We use 10 semantic classes (one free-space, nine occupied), merging rare categories: bicycles, motorcycles, and scooters into two-wheeler, and several static structural categories into other-structure.

\subsection{Baseline Implementations}

\subsubsection{Monocular BEV-based baselines.}
Monocular baselines include BEVDet4D~\cite{huang2022bevdet4d} as a detection and velocity baseline, and FlashOcc~\cite{yu2023flashocc} and Panoptic-FlashOcc~\cite{yu2024panoptic} as semantic and panoptic 3D occupancy baselines. We use 8 historical frames based on their public settings. The original BEVDet4D, FlashOcc, and Panoptic-FlashOcc models assume multi-view surround-camera inputs, whereas our dataset contains only front-view images. Therefore, we adapt them to a monocular setting by using only the left stereo camera to estimate depth. We also introduce Panoptic-FlashOcc-vel, which extends the detection head of Panoptic-FlashOcc to predict velocities $(v_x, v_y)$ with an L1 loss for velocity supervision. 

\subsubsection{Stereo baseline: VoxFormer\textnormal{~\cite{li2023voxformer}}.} We adapt VoxFormer-T with 4 historical frames to our stereo setup, sampling frames alternately in time to match the temporal coverage of the 8-frame monocular baselines. Stage-1 takes voxelized stereo-matching depth as input and predicts a 3D occupancy volume at 0.4m resolution, which is then upsampled to our 0.2m target grid. We follow the public training settings of each baseline.

\subsection{Benchmark Results}
\subsubsection{3D occupancy prediction.} \cref{tab:occ_comparison} and~\cref{fig:occ_comparison} compare the baselines on the semantic occupancy prediction task. FlashOcc (FO) achieves the highest mean IoU (mIoU) and outperforms the other methods on most classes. Panoptic-FlashOcc (PF) yields a minor mIoU drop relative to FO due to the multi-task trade-off. VoxFormer (VF) remains competitive on pedestrians, possibly because it benefits from stereo depth, whereas FlashOcc may inherit monocular pedestrian detection errors from BEVDet4D pretraining. Cars and trucks show lower IoU than in self-driving benchmarks, primarily because they are less common in our dataset.

\subsubsection{Panoptic occupancy.} We further evaluate the pedestrian detection and panoptic occupancy quality using BEVDet4D and Panoptic-FlashOcc. As shown in~\cref{tab:panoptic_vel} and~\cref{fig:pano_comparison}, BEVDet4D (BD) provides only pedestrian detections, while Panoptic-FlashOcc (PF) predicts voxel-level panoptic labels. Pedestrian Average Precision ({AP$^{\rm Ped}$}) is evaluated across distance thresholds 0.1\,m, 0.2\,m, 0.5\,m and 1\,m on detected centers. In addition to standard panoptic metrics, we also report PQ$^{\dagger}$, a relaxed panoptic quality~\cite{porzi2019seamless} that relaxes the strict IoU requirement for stuff classes and is better suited for vision-only occupancy prediction.

\subsubsection{Pedestrian velocity prediction.} Finally, we compare BEVDet4D and Panoptic-FlashOcc-vel (PF-vel) on the pedestrian velocity prediction task. \cref{tab:panoptic_vel} summarizes the absolute velocity error (AVE; m/s) over all ground truth pedestrian voxels (AVE-T), true-positive detections within 1\,m (AVE-D), and correctly classified voxels occupied by pedestrians (AVE-O), together with mIoU. PF-vel achieves comparable velocity accuracy while maintaining voxel-level semantic occupancy. Qualitative examples in~\cref{fig:flow_comparison} show that PF-vel produces plausible velocities for most pedestrians but still exhibits typical failure cases, such as confusion between forward and backward walking directions, indicating that there is remaining headroom for improving pedestrian velocity prediction in conjunction with panoptic occupancy.

\begin{table}[t]
\centering
\scriptsize
\setlength{\tabcolsep}{2pt}
\caption{3D occupancy prediction performance. VF-T: VoxFormer-T, FO: FlashOcc, PF: Panoptic-FlashOcc. Best in \textbf{bold}. IoU denotes geometric IoU.}
\label{tab:occ_comparison}
\begin{tabular}{@{}l!{\vrule width \lightrulewidth}c>{\columncolor{mgray}}c!{\vrule width \lightrulewidth}ccccccccc@{}}
\toprule
Method & IoU & mIoU & \clsicon{pedestrian} & \clsicon{car} & \clsicon{otherstruct} & \clsicon{pole} & \clsicon{road} & \clsicon{terrain} & \clsicon{truck} & \clsicon{twowheeler} & \clsicon{vegetation} \\
\midrule
VF-T    & 57.81 & 24.89 & \textbf{32.79} & 1.39 & 28.47 & 7.08 & \textbf{70.90} & 23.57 & 5.23 & 28.04 & 26.52 \\
FO (8f) & 57.71 & \textbf{27.18} & 31.91 & \textbf{5.36} & 31.00 & \textbf{10.16} & 70.82 & \textbf{27.29} & \textbf{7.35} & 30.85 & \textbf{29.88} \\
PF (8f) & \textbf{57.83} & 26.64 & 32.45 & 3.66 & \textbf{31.79} & 9.96 & 70.35 & 25.90 & 5.87 & \textbf{30.99} & 28.83 \\
\bottomrule
\end{tabular}\\[3pt]
{\scriptsize
\clsicon{pedestrian}\,ped.\quad
\clsicon{car}\,car\quad
\clsicon{otherstruct}\,oth.str.\quad
\clsicon{pole}\,pole\quad
\clsicon{road}\,road\quad
\clsicon{terrain}\,terrain\quad
\clsicon{truck}\,truck\quad
\clsicon{twowheeler}\,2-wh.\quad
\clsicon{vegetation}\,veg.}
\end{table}

\begin{table}[t]
\centering
\scriptsize
\setlength{\tabcolsep}{1.5pt}
\caption{Panoptic occupancy and pedestrian velocity prediction. BD: BEVDet4D, PF: Panoptic-FlashOcc, PF-vel: Panoptic-FlashOcc-vel.}
\label{tab:panoptic_vel}
\begin{tabular}{@{}l!{\vrule width \lightrulewidth}cccc!{\vrule width \lightrulewidth}ccc!{\vrule width \lightrulewidth}c!{\vrule width \lightrulewidth}ccc!{\vrule width \lightrulewidth}c@{}}
\toprule
& \multicolumn{4}{c!{\vrule width \lightrulewidth}}{Scene-level} & \multicolumn{3}{c!{\vrule width \lightrulewidth}}{Pedestrian} & & \multicolumn{3}{c!{\vrule width \lightrulewidth}}{AVE (m/s)$\downarrow$} & \\
\cmidrule(lr){2-5} \cmidrule(lr){6-8} \cmidrule(lr){10-12}
Method & PQ & PQ$^\dagger$ & RQ & SQ & PQ$^\text{Ped}$ & RQ$^\text{Ped}$ & SQ$^\text{Ped}$ & AP$^\text{Ped}$ & T & D & O & mIoU$\uparrow$ \\
\midrule
BD (8f)     & --   & --   & --   & --   & --   & --   & --   & 41.7 & 1.00 & 0.36 & --   & --    \\
PF (8f)     & 19.9 & 32.6 & 28.1 & 65.8  & 42.5 & 60.2 & 70.7 & 45.5 & --   & --   & --   & --    \\
PF-vel (8f) & --   & --   & --   & --   & --   & --   & --   & --   & 0.97 & 0.39 & 0.67 & 26.00 \\
\bottomrule
\end{tabular}
\end{table}
\section{Conclusion}
\label{sec:conclusion}

This paper presents MobileOcc, a human-aware semantic occupancy dataset for mobile robots operating in pedestrian-dense, near-field environments. We provide baselines for occupancy prediction and pedestrian-velocity estimation across monocular, stereo, and panoptic settings. We further validate the human-mesh optimization component on established datasets, where adding LiDAR points for optimization substantially improves image-based methods and remains robust across them. Together, the dataset, pipeline, and benchmarks form a practical testbed for perception stacks that must jointly reason about humans, static objects, and free space to enable safe, precise navigation in crowds.

\subsubsection*{Limitations and future work.} Currently, MobileOcc focuses on outdoor scenes, which may limit generalization across domains and conditions. Future extensions include broadening geographic and environmental diversity (e.g., indoor spaces and adverse weather). Additionally, our non-rigid model uses SMPL and does not capture carried or accompanying objects; we plan to extend the benchmark toward human-object occupancy to better handle such cases.

\section*{Acknowledgements}
This work was supported by the European Union through ERC INTERACT, Grant 101041863. Views and opinions expressed are, however, those of the author(s) only and do not necessarily reflect those of the European Union. Neither the European Union nor the granting authority can be held responsible for them.
{
    \small
    \bibliographystyle{ieeenat_fullname}
    \bibliography{main}
}
\clearpage
\setcounter{page}{1}
\maketitlesupplementary

\appendix
\setcounter{section}{0}
\renewcommand{\thesection}{\Alph{section}}
\renewcommand{\thesubsection}{\Alph{section}.\arabic{subsection}}

\section{Visibility filtering}
\label{sec:visibility}

We follow the notation introduced in Sec.~\ref{subsec:smpl_dynamic} of the main paper:
SMPL parameters $(\boldsymbol{\beta},\boldsymbol{\theta},\mathbf{t}_{\mathrm{cam}})$,
mesh $\mathcal{M}$ with visible subset $\mathcal{V}\subset\mathcal{M}$, LiDAR point
cloud $\mathcal{P}$, and camera intrinsics $\mathbf{K}$. Our visibility filtering selects a subset of mesh vertices that i) are geometrically visible from the camera and ii) belong to body parts that are likely unoccluded in the image. The resulting visible subset $\mathcal{V}$ is then used for ICP registration and for the 3D LiDAR alignment loss $\mathcal{L}_{3D}$.

\subsection{Backface culling}
The SMPL mesh $\mathcal{M}$ consists of triangular faces, each with three vertices in 3D.
For each face $f$, we compute i) a unit face normal $\mathbf{n}_f$ that indicates which direction the face is pointing, and ii) a representative point $\mathbf{c}_f$ at the center (centroid) of the triangle. Let $\mathbf{o}$ denote the 3D position of the camera. The viewing direction from the camera to face $f$ is then
\begin{equation}
    \mathbf{p}_f =
    \frac{\mathbf{c}_f - \mathbf{o}}
         {\left\lVert \mathbf{c}_f - \mathbf{o} \right\rVert_2}.
\end{equation}

We classify a face as front-facing if the angle between its normal and the viewing direction is sufficiently small. We implement this using the inner product between the normal and the viewing direction:
\begin{equation}
    \mathbf{n}_f^\top \mathbf{p}_f < w,
\end{equation}
where $w$ is a backface-culling threshold (tuned per dataset, see Table~\ref{tab:scalar-weights}). All faces that satisfy this condition are considered front-facing, and all vertices that belong to at least one such face form a candidate visible set $\mathcal{V}_{\text{front}}\subset\mathcal{M}$.

This step removes mesh regions oriented away from the camera. It therefore should not be matched to LiDAR points, thereby reducing the influence of faces that are clearly not visible from the current viewpoint.

\subsection{Body-part filtering from 2D keypoints}
Backface culling alone does not handle self-occlusion (e.g., the arm on the far side of the body) or occlusion by other objects. To further restrict the mesh to regions observed, we use the same 2D keypoints and confidence scores as in the 2D joint reprojection loss, $\mathcal{L}_J$.

Let $\mathbf{J}^{\mathrm{est}}_i$ denote the detected 2D joints and $w_i$ their confidence scores. For each SMPL body part $b$ (e.g., torso, left upper arm, right lower leg), we define a small set of 2D joints $\mathcal{J}_b$ that are informative for that part (for example, shoulder and elbow for an upper arm). A body part is marked as occluded if all of its associated joints have low
confidence:
\begin{equation}
    \max_{i \in \mathcal{J}_b} w_i < J_{\text{conf}},
\end{equation}
where $J_{\text{conf}}$ is a dataset-specific threshold, listed in Table~\ref{tab:scalar-weights}.

For each occluded body part, we remove all mesh faces whose vertices belong to that part from the candidate visible set $\mathcal{V}_{\text{front}}$ obtained by backface culling. Since the SMPL template provides a fixed correspondence between mesh faces and semantic body parts, this mapping is known in advance and does not depend on pose or shape. The remaining faces define a refined visible mesh that is consistent with both the camera viewpoint and the 2D keypoint evidence. This body-part filtering improves robustness in cases with occluded limbs by preventing LiDAR points from being matched to mesh regions that are not supported by the image.

\subsection{Final visible vertex set}
Combining backface culling and body-part filtering yields our final set of
visible mesh vertices $\mathcal{V} \subset \mathcal{M}$. By construction,
\begin{itemize}
    \item $\mathcal{V}$ contains only vertices that belong to front-facing faces, and
    \item vertices assigned to body parts marked as occluded by the 2D keypoints
          are removed.
\end{itemize}
We treat $\mathcal{V}$ as the subset of the mesh that is both visible in the
image and observable by the LiDAR sensor. All subsequent 3D processing is
restricted to $\mathcal{V}$. Focusing on $\mathcal{V}$ instead of the full mesh
reduces the influence of self-occluded or truncated limbs and leads to more
stable LiDAR-mesh alignment in scenes with occlusions.

\section{Scalar Weights and Optimization}
\label{sec:weights}

The mesh refinement in Section~\ref{subsec:smpl_dynamic} of the main paper minimizes the multi-term
objective in Eq.~\ref{eq:total}, combining 2D joint reprojection $\mathcal{L}_{J}$, 3D LiDAR alignment
$\mathcal{L}_{3D}$, pose and shape priors $\mathcal{L}_{\theta}$ and $\mathcal{L}_{\beta}$, an
anti-hyperextension prior $\mathcal{L}_{a}$, and an occlusion-aware pose consistency term
$\mathcal{L}_{\theta,\mathrm{occ}}$. All terms are fully differentiable with respect to\
$(\boldsymbol{\beta},\boldsymbol{\theta},\mathbf{t}_{\mathrm{cam}})$.

Our optimization scheme contains several scalar hyperparameters:
\begin{itemize}
    \item Loss weights $\lambda_{\theta}$, $\lambda_{a}$, $\lambda_{\beta}$, $\lambda_{3D}$, and $\lambda_{\text{occ}}$.
    \item The Geman--McClure scale $\rho$ used in $L_J$.
    \item The backface culling threshold $w$ used in the visibility filter.
    \item The joint confidence threshold $J_{\text{conf}}$ used for part-level occlusion reasoning.
\end{itemize}
These hyperparameters influence the trade-off between image alignment, LiDAR alignment, and prior regularization. 
To obtain the best configurations, we perform Bayesian Optimization on a subset that is held out from the final evaluation for each dataset. For each trial, we run the full optimization pipeline with a candidate parameter configuration and evaluate the resulting meshes using the relevant 3D metrics (e.g., PVE, MPJPE, PA-MPJPE). Bayesian Optimization maintains a surrogate model of the objective over the hyperparameter space and iteratively proposes new configurations that balance exploration and exploitation. In practice, we run 100 iterations per dataset and select the best-performing configuration on the subset.

Table~\ref{tab:scalar-weights} lists the resulting values for all datasets used in our experiments. Note that the UT Campus dataset does not provide ground-truth SMPL annotations; for this dataset, we select hyperparameters based on qualitative inspection of the resulting meshes.

\begin{table}[h]
    \centering
    \footnotesize
    \setlength{\tabcolsep}{4pt}
    \begin{tabular}{lcccccccc}
        \toprule
        Dataset & $\rho$ & $\lambda_{\theta}$ & $\lambda_{a}$ & $\lambda_{\beta}$ & $\lambda_{3D}$ & $\lambda_{\text{occ}}$ & $w$ & $J_{\text{conf}}$ \\
        \midrule
        3DPW        & 100 & 2.2  & 11.0 & 5.0  & 800 & 35  & 0.2 & 0.6 \\
        SLOPER4D    & 100 & 0.75 & 8.5 & 1.0  & 500 & 55  & 0.2 & 0.7 \\
        HumanM3     & 100 & 1.0  & 3.0 & 17.5 & 600 & 135 & $-1.0$ & 0.6 \\
        UT Campus    & 100 & 1.4  & 10.0 & 10.0 & 300 & 50  & 0.2 & 0.7 \\
        \bottomrule
    \end{tabular}
    \caption{Sub-optimal scalar weights and thresholds per dataset.}
    \label{tab:scalar-weights}
\end{table}

In all experiments, we start from the initial mesh prediction provided by the base HMR model and run gradient descent on $(\beta,\theta,t_{\text{cam}})$ until convergence or a fixed iteration budget is reached. The chosen hyperparameters ensure that the optimization improves both image and LiDAR alignment while keeping poses and shapes within a plausible region of the SMPL space and preventing excessive drift of occluded limbs.

\section{Synthetic LiDAR sweeps}
\label{sec:synthetic-lidar}

For datasets without real LiDAR (3DPW~\cite{von2018recovering}), we simulate LiDAR sweeps directly on the ground-truth SMPL meshes to study how LiDAR characteristics affect our method. For each camera frame, we place a virtual LiDAR at the camera center and emit rays based on the vertical and horizontal angular resolutions of an Ouster-style spinning sensor. Each ray intersects the SMPL mesh to obtain a 3D return, analogous to a time-of-flight measurement. If a ray does not hit the mesh within the valid range, it produces no return.

We model three virtual sensors with increasing angular resolutions: ``Ouster-32'', ``Ouster-64'', and ``Ouster-128''. For each simulated return, we add range and angular perturbations and randomly drop individual points to mimic missed detections. All rays are restricted to the camera frustum, so the simulated LiDAR only observes the same region as the RGB camera. Since 3DPW does not contain full 3D scene geometry, occlusions from other objects are not simulated. Instead, our visibility filter (Section~\ref{sec:visibility}) discards mesh regions that are likely occluded in the image.

The sensor configurations are summarized in Table~\ref{tab:lidar-specs}. All three sensors share the same noise characteristics and valid distance range $[0.5,90]$\,m, but differ in vertical and horizontal resolution. We use these simulated sweeps as input to our pipeline, keeping all other components identical to the real-LiDAR setting. This design allows us to isolate the impact of LiDAR density and noise on performance.

\begin{table}[h]
    \centering
    \footnotesize
    \setlength{\tabcolsep}{3pt}
    \begin{tabular}{l l cc cc cc cc}
        \toprule
        \textit{Nr.} & \textit{Type} &
        \multicolumn{2}{c}{Resolution} &
        \multicolumn{2}{c}{Range noise} &
        \multicolumn{2}{c}{Angular noise} &
        \multicolumn{2}{c}{Range} \\
        \cmidrule(lr){3-4} \cmidrule(lr){5-6} \cmidrule(lr){7-8} \cmidrule(lr){9-10}
        & &
        Vert. & Hor. &
        mean & std &
        mean & std &
        min & max \\
        \midrule
        1. & Ouster-32  &  32 &  512  & $\pm 25.0$ & 10.0 & 0.0 & 0.01 & 0.5 & 90.0 \\
        2. & Ouster-64  &  64 & 1024  & $\pm 25.0$ & 10.0 & 0.0 & 0.01 & 0.5 & 90.0 \\
        3. & Ouster-128 & 128 & 2048  & $\pm 25.0$ & 10.0 & 0.0 & 0.01 & 0.5 & 90.0 \\
        \bottomrule
    \end{tabular}
    \caption{Specifications of the simulated LiDAR sensors used on 3DPW~\cite{von2018recovering}.
    Resolution is given in vertical and horizontal channels, range noise mean/std are in millimeters, angular noise mean/std are in degrees, and range min/max are in meters. All sensors use the same noise levels and a fixed dropout probability of 10\%.}
    \label{tab:lidar-specs}
\end{table}

The simulated LiDAR sweeps approximate realistic Ouster sensors in terms of angular
resolution, range noise, and dropout behavior, while being perfectly time-synchronized with the
RGB images and SMPL ground truth. This provides a controlled setting to quantify how much
additional 3D information is needed for our optimization to improve over purely image-based HMR.

\begin{table}[h]
    \centering
    \footnotesize
    \setlength{\tabcolsep}{3pt}
    \begin{tabular}{l l cc cccc}
        \toprule
        & & \multicolumn{2}{c}{Ours} & \multicolumn{4}{c}{No visibility filtering} \\
        \cmidrule(lr){3-4} \cmidrule(lr){5-8}
        Sensor & Metric & mean & std & mean & std & t-value & p-value \\
        \midrule
        \multirow{3}{*}{Ouster-32} 
            & PVE $\downarrow$      & 82.0 & 50.9 & 84.3 & 60.2 &  9.512 & 0.3416 \\
            & MPJPE $\downarrow$    & 62.9 & 42.6 & 64.9 & 52.7 &  9.623 & 0.3360 \\
            & PA-MPJPE $\downarrow$ & 50.2 & 35.0 & 51.5 & 42.3 &  7.720 & 0.4402 \\
        \midrule
        \multirow{3}{*}{Ouster-64} 
            & PVE $\downarrow$      & 66.9 & 40.0 & 69.3 & 47.6 & 12.585 & 0.2082 \\
            & MPJPE $\downarrow$    & 48.2 & 31.0 & 50.4 & 36.4 & 15.002 & 0.1337 \\
            & PA-MPJPE $\downarrow$ & 41.1 & 29.4 & 42.3 & 34.8 &  8.588 & 0.3905 \\
        \midrule
        \multirow{3}{*}{Ouster-128} 
            & PVE $\downarrow$      & 58.7 & 35.4 & 60.2 & 42.2 &  8.879 & 0.3747 \\
            & MPJPE $\downarrow$    & 43.6 & 26.4 & 45.3 & 35.6 & 12.506 & 0.2112 \\
            & PA-MPJPE $\downarrow$ & 37.2 & 27.1 & 38.7 & 33.5 & 11.350 & 0.2565 \\
        \bottomrule
    \end{tabular}
    \caption{Ablation study on the visibility filter for partially occluded body parts. ``Ours'' denotes the full model with the visibility filter, while ``No visibility filtering'' disables this component.}
    \label{tab:ablation-visibility}
\end{table}

\subsection{Ablation study}
\label{subsec:ablation}

\begin{table*}[h]
    \centering
    \footnotesize
    \setlength{\tabcolsep}{7pt}
    \begin{tabular}{l l cc cccc cccc}
        \toprule
        & & \multicolumn{2}{c}{Ours} &
        \multicolumn{4}{c}{No $\mathcal{L}_{\theta,\mathrm{occ}}$} &
        \multicolumn{4}{c}{No $\mathcal{L}_{3D}$} \\
        \cmidrule(lr){3-4} \cmidrule(lr){5-8} \cmidrule(lr){9-12}
        Sensor & Metric &
        mean & std &
        mean & std & t-value & p-value &
        mean & std & t-value & p-value \\
        \midrule
        \multirow{3}{*}{Ouster-32}
            & PVE $\downarrow$      & 73.2 & 50.9 &  75.9 & 75.9 & -68.66 & $0.00$          & 171.2 & 113.1 & -191.95 & $0.00$ \\
            & MPJPE $\downarrow$    & 57.0 & 42.6 &  58.2 & 43.2 & -66.92 & $0.00$          & 159.3 & 113.2 & -194.86 & $0.00$ \\
            & PA-MPJPE $\downarrow$ & 47.6 & 35.0 &  48.9 & 35.5 & -53.33 & $0.00$          &  49.6 &  22.1 & -78.44  & $0.00$ \\
        \midrule
        \multirow{3}{*}{Ouster-64}
            & PVE $\downarrow$      & 57.2 & 40.0 &  60.1 & 41.3 & -33.20 & $1.09\times 10^{-23}$ & 173.2 & 117.4 & -188.63 & $0.00$ \\
            & MPJPE $\downarrow$    & 43.9 & 31.0 &  45.2 & 31.8 & -27.73 & $0.00$          & 161.7 & 117.8 & -191.47 & $0.00$ \\
            & PA-MPJPE $\downarrow$ & 38.5 & 29.4 &  40.4 & 31.0 & -24.21 & $0.00$          &  49.2 &  21.6 & -77.00  & $0.00$ \\
        \midrule
        \multirow{3}{*}{Ouster-128}
            & PVE $\downarrow$      & 50.5 & 35.4 &  52.4 & 36.8 &  -7.18 & $1.43\times 10^{-12}$ & 174.9 & 121.4 & -185.41 & $0.00$ \\
            & MPJPE $\downarrow$    & 39.1 & 26.4 &  40.3 & 27.6 &  -5.74 & $9.71\times 10^{-9}$  & 163.4 & 121.8 & -188.04 & $0.00$ \\
            & PA-MPJPE $\downarrow$ & 35.1 & 27.1 &  36.6 & 28.9 &  -7.27 & $3.62\times 10^{-13}$ &  49.4 &  23.0 & -76.71  & $0.00$ \\
        \bottomrule
    \end{tabular}
    \caption{Ablation study on the proposed loss terms in the objective function (Eq.~\ref{eq:total}), evaluated on the full 3DPW~\cite{von2018recovering} test set (35{,}515 frames). Removing the occlusion-aware pose prior $\mathcal{L}_{\theta,\mathrm{occ}}$ slightly degrades performance, while removing the 3D LiDAR alignment term $\mathcal{L}_{3D}$ leads to a drastic increase in error across all sensors.}
    \label{tab:ablation-loss}
\end{table*}

We perform an ablation study on 3DPW~\cite{von2018recovering} to quantify the contribution of each newly introduced component: i) the visibility filter, ii) the 3D LiDAR alignment term $\mathcal{L}_{3D}$, and iii) the occlusion-aware pose prior $\mathcal{L}_{\theta,\mathrm{occ}}$. For each of the three simulated sensors, we run our method with one component removed at a time and compare PVE, MPJPE, and PA-MPJPE to the full model.

Table~\ref{tab:ablation-visibility} focuses on the visibility filter. Since this component only affects frames where at least one body part is (partially) occluded, we construct a subset of 1,063 frames from 3DPW~\cite{von2018recovering} that each contain at least one occluded limb (about 2.9\% of the dataset). On this subset, we compare our full model with visibility filtering to a variant where the filter is disabled. Across all three simulated sensors, removing the visibility filter consistently increases PVE, MPJPE, and PA-MPJPE, although the numerical differences are modest. This confirms that masking out occluded regions helps in precisely those challenging cases, while having little effect on fully visible poses.

Table~\ref{tab:ablation-loss} investigates the two new loss terms from Section~\ref{subsec:smpl_dynamic}: the occlusion-aware pose prior $\mathcal{L}_{\theta,\mathrm{occ}}$ and the 3D LiDAR alignment term $\mathcal{L}_{3D}$. Here, we evaluate on the full 3DPW test set (35,515 frames). For each sensor, we report the performance of the full model (``Ours'') and compare it to i) a variant without $\mathcal{L}_{\theta,\mathrm{occ}}$ and ii) a variant without $\mathcal{L}_{3D}$.

The 3D term $\mathcal{L}_{3D}$ has by far the most significant impact. Removing it roughly doubles the PVE and MPJPE across all three sensors, bringing performance close to that of using only ICP alignment without any LiDAR-aware refinement. This confirms that the LiDAR–mesh Chamfer term is crucial for correcting local pose and shape errors beyond what ICP alone can provide.

The occlusion-aware pose prior $\mathcal{L}_{\theta,\mathrm{occ}}$ yields a more minor but consistent gain: disabling this term slightly worsens all metrics, indicating that softly constraining occluded joints toward their initial configuration helps prevent unrealistic drift while still allowing visible joints to move freely.

\section{Baseline Details}
\label{sec:baseline-supp}
\textbf{BEVDet4D.}
The public configuration uses an effective batch size of 64 (8×8), a learning rate of $2\times10^{-4}$, 
and 20 epochs. We retain these settings but train on a single A40 using batch~8 with 8-step gradient 
accumulation, and shorten training to 15 epochs. The BEV and depth ranges are 
$x\!\in[0,12.8]$\,m, $y\!\in[-6.4,6.4]$\,m, and depth $[1,15]$\,m with 0.5\,m bins.

\textbf{FlashOcc and its variants.}
All variants are initialized from our BEVDet4D~(1f) checkpoint (12 epochs). The original setup uses 
LR~$1\times10^{-4}$, effective batch~16 (4×4), and 24 epochs; we match the batch size on a single A40 
via batch~4 with 4-step accumulation and train for 15 epochs. The occupancy volume follows our final 
evaluation grid: $x\!\in[0.4,10.0]$\,m, $y\!\in[-4.8,4.8]$\,m, $z\!\in[-1.0,3.8]$\,m at 0.2\,m 
resolution, with depth bins $[1,12]$\,m at 0.5\,m.

\textbf{VoxFormer.} For VoxFormer-T we keep the public training settings (20 epochs and learning rate $2\times 10^{-4}$). The original configuration uses batch size 1 on 8 GPUs (effective batch 8); on our single A40 we use batch size 1 with 8-step gradient accumulation to match this.

\textbf{Input processing and augmentation.}
All methods use the same image pipeline: input images of $1024\times 1224$ (H$\times$W) are vertically cropped from rows 144 to 1008 (yielding $864\times 1224$) and then resized to $384\times 544$. BEVDet4D and the FlashOcc family use the default image augmentations from their public code (no BEV-space augmentation). For fairness, we also implement a random horizontal flip for VoxFormer in the image space.

\textbf{EMA and checkpoints.} The public VoxFormer code does not maintain an exponential moving average (EMA) model, so we report results from the final checkpoint. BEVDet4D and the FlashOcc family use EMA in their official implementations, and we therefore evaluate these baselines using their EMA weights.

\end{document}